\documentclass[runningheads,a4paper]{llncs}

\usepackage{amssymb}
\usepackage{graphicx}

\usepackage{url}
\urldef{\mailsa}\path|francisco.rangel@autoritas.es|
\urldef{\mailsb}\path|mfranco@prhlt.upv.es|
\urldef{\mailsc}\path|prosso@dsic.upv.es|    
\newcommand{\keywords}[1]{\par\addvspace\baselineskip
\noindent\keywordname\enspace\ignorespaces#1}

\usepackage{times}
\usepackage{latexsym}
\usepackage{graphicx}
\usepackage{amsfonts}
\usepackage{amsmath}
\usepackage{url}
\usepackage{color}
\usepackage{multirow}

\usepackage{float}
\usepackage{hhline}
\usepackage{paralist}

\usepackage{booktabs}

\begin{document}

\mainmatter  

\title{A Low Dimensionality Representation\\for  Language Variety Identification\thanks{The work of the first author was in the framework of ECOPORTUNITY IPT-2012-1220-430000. The work of the last two authors was in the framework of the SomEMBED MINECO TIN2015-71147-C2-1-P research project. This work has been also supported by the SomEMBED TIN2015-71147-C2-1-P MINECO research project and by the Generalitat Valenciana under the grant ALMAPATER (PrometeoII/2014/030)}}

\titlerunning{A Low Dimensionality Representation\\for  Language Variety Identification}

%
%
\author{Francisco Rangel$^{1,2}$ \and Marc Franco-Salvador$^1$ \and Paolo Rosso$^1$}
\authorrunning{Francisco Rangel \and Marc Franco-Salvador \and Paolo Rosso}

\institute{$^1$Universitat Polit\`{e}cnica de Val\`{e}ncia, Spain\\
$^2$Autoritas Consulting, Spain\\
\mailsa, 
\mailsb,
\mailsc\\
}

%
%

\toctitle{A Low Dimensionality Representation}
\tocauthor{for Language Variety Identification}
\maketitle

\begin{abstract}
Language variety identification aims at labelling texts in a native language (e.g. Spanish, Portuguese, English) with its specific variation (e.g. Argentina, Chile, Mexico, Peru, 
Spain; Brazil, Portugal; UK, US). In this work we propose a low dimensionality representation (LDR) to address this task with five different varieties of Spanish: Argentina, 
Chile, 
Mexico, Peru and Spain. We compare our LDR method with common state-of-the-art representations and show an increase in accuracy of $\sim$35\%. Furthermore, we compare LDR with two reference
distributed representation models. Experimental results show competitive performance while dramatically reducing the dimensionality --- and increasing the big data suitability --- 
to only 6 features per variety. Additionally, we analyse the behaviour of the employed machine learning algorithms and 
the most discriminating features. Finally, we employ an alternative dataset to test the robustness of our low dimensionality representation with another set of similar languages.

\end{abstract}

\section{Introduction}
\label{intro}

Language variety identification aims at labelling texts in a native language (e.g. Spanish, Portuguese, English) with their specific variation (e.g. Argentina, Chile, Mexico, 
Peru, 
Spain; Brazil, Portugal; UK, US). Although at first sight language variety identification may seem a classical text classification problem, cultural idiosyncrasies may influence 
the way users construct their discourse, the kind of sentences they build, the expressions they use or their particular choice of words. Due to that, we can consider language 
variety identification as a double problem of text classification and author profiling, where information about how language is shared by people may help to discriminate among 
classes of authors depending on their language variety.

This task is specially important in social media. Despite the vastness and accessibility of the Internet destroyed frontiers among regions or traits, companies are still very 
interested in author profiling segmentation. For example, when a new product is launched to the market, knowing the geographical distribution of opinions may help to improve 
marketing campaigns. Or given a security threat, knowing the possible cultural idiosyncrasies of the author may help to better understand who could have written the message. 

Language variety identification is a popular research topic of natural language processing. In the last years, several tasks and workshops have been organized: the Workshop on 
Language Technology for Closely Related Languages and Language Variants @ EMNLP 2014\footnote{\url{http://alt.qcri.org/LT4CloseLang/index.html}}; the VarDial Workshop @ COLING 
2014 - Applying NLP Tools to Similar Languages, Varieties and Dialects\footnote{\url{http://corporavm.uni-koeln.de/vardial/sharedtask.html}}; and the LT4VarDial - Joint Workshop 
on Language Technology for Closely Related Languages, Varieties and Dialect\footnote{\url{http://ttg.uni-saarland.de/lt4vardial2015/dsl.html}} @ RANLP 
\cite{zampieri2014}\cite{tan2014}. 
We can find also several works focused on the task.
In~\cite{sadat} the authors addressed the problem of identifying Arabic varieties in blogs and social fora. They used character $n$-gram features 
to discriminate between six different varieties and obtained accuracies between 70\%-80\%. Similarly, 
\cite{zampieri} collected 1,000 news articles of two varieties of Portuguese. They applied different features such as word and character 
$n$-grams and reported accuracies over 90\%. With respect to the Spanish 
language, \cite{maier2014} focused on varieties from Argentina, Chile, Colombia, Mexico and Spain in Twitter. They used meta-learning and combined four types of 
features: \textit{i)} character $n$-gram frequency profiles, \textit{ii)} character $n$-gram language models, \textit{iii)} Lempel-Ziv-Welch compression and \textit{iv)} 
syllable-based language models. They obtained an interesting 60\%-70\% accuracy of classification.

We are interested in discovering which kind of features capture higher differences among varieties. Our hypothesis is that language varieties differ mainly in lexicographic 
clues. We show an example in Table \ref{tab:sentences}.

\vspace{-0.5cm}
\begin{table}[H]
\begin{center}
\begin{tabular}{|l| l l l | }

    \hline
   \textbf{English}$~~~~~~~~~~~~~~$ &     \multicolumn{3}{p{8cm}|}{
   	I was \textbf{goofing around} with my dog and I \textbf{lost} my \textbf{mobile}.
  }\\
     \hline
\textbf{ES-Argentina}$~~~~~~~~~~~~~~$ &     \multicolumn{3}{p{8cm}|}{
   	Estaba haciendo \textbf{boludeces} con mi perro y \textbf{extravi\'e} el \textbf{celular}.
  }\\
       \hline
\textbf{ES-Mexico}$~~~~~~~~~~~~~~$ &     \multicolumn{3}{p{8cm}|}{
   	Estaba haciendo el \textbf{pendejo} con mi perro y \textbf{extravi\'e} el \textbf{celular}.
  }\\
       \hline
\textbf{ES-Spain}$~~~~~~~~~~~~~~$ &     \multicolumn{3}{p{8cm}|}{
   	Estaba haciendo el \textbf{tonto} con mi perro y \textbf{perd\'i} el \textbf{m\'ovil}.
  }\\
       \hline
\end{tabular}
\end{center}
\caption{The same example in three varieties of Spanish (Argentina, Mexico and Spain).}
\label{tab:sentences}
\end{table}

\vspace{-0.5cm}
In this work we focus on the Spanish language variety identification. We differentiate from the previous works as follows: \textit{i)} instead of $n$-gram based 
representations, we propose a low dimensionality representation that is helpful when dealing with big data in social media; \textit{ii)} in order to reduce the possible 
over-fitting, our training and test partitions do not share any author of instance between them\footnote{It is important to highlight the importance of this aspect from an 
evaluation perspective in an author profiling scenario. In fact, if texts from the same authors are both part of the training and test sets, their particular style and vocabulary 
choice may contribute at training time to learn the profile of the authors. In consequence, over-fitting would be biasing the results.}; and \textit{iii)} in contrast to the 
Twitter dataset of \cite{maier2014}, we will make available our dataset to the research community. 

\section{Low Dimensionality Representation}

The key aspect of the low dimensionality representation (LDR) is the use of weights to represent the probability of each term to belong to each one of the different language 
varieties. We assume that the distribution of weights for a given document should be closer to the weights of its corresponding language variety. Formally, the LDR is estimated as 
follows:

\paragraph{\textbf{Term-frequency - inverse document frequency (tf-idf) matrix creation.}} First, we apply the \textit{tf-idf}~\cite{salton1988term} weighting for the terms of the 
documents of the training set \textit{D}. As result we obtain the following matrix:

\begin{equation}
\Delta=\begin{bmatrix}
w_{11}&w_{12}&...&w_{1m} & \delta(d_1) \\
w_{21}&w_{22}&...&w_{2m} & \delta(d_2)\\
...&...&...&...& \\
w_{n1}&w_{n2}&...&w_{nm} & \delta(d_n)\\
\end{bmatrix},
\label{eq:2}
\end{equation}

\noindent where each row in the matrix \begin{math} \Delta \end{math} represents a document \textit{d}, each column represents a vocabulary term \textit{t}, 
\begin{math}w_{ij}\end{math} represents its \textit{tf-idf}, and \begin{math}\delta(d_i)\end{math} represents the assigned class \textit{c} of the 
document \textit{i}, that is, the language variety actually assigned to this document. 

\paragraph{\textbf{Class-dependent term weighting.}} Using the matrix $\Delta$, we obtain the class-dependent term weight matrix $\beta$. This matrix contains the weights of each 
term $t$ for each language variety $C$ on the basis of Eq.~\ref{eq:3}:

\begin{equation}
W(t,c) = \frac{\sum_{d\in{D}/c=\delta(d)}w_{dt}}{\sum_{d\in{D}}w_{dt}}, \forall{d\in{D}, c\in{C}}
\label{eq:3}
\end{equation}

Basically, the term weight $W(t,c)$ is the ratio between the weights of the documents belonging to a concrete language variety $c$ and the total distribution of weights for that 
term $t$.

\paragraph{\textbf{Class-dependent document representation.}} We employ the class-dependent term weights $\beta$ to obtain the final representation of the documents as follows:

\begin{equation}
d = \{F(c_1), F(c_2), ..., F(c_n)\} \sim \forall{c\in{C}},
\label{eq:4}
\end{equation}

\begin{equation}
F(c_i) = \{avg, std, min, max, prob, prop\} 
\label{eq:5}
\end{equation}

\noindent where each \begin{math}F(c_i)\end{math} contains the set of features showed in Eq.~\ref{eq:5} and described in Table~\ref{tab:features}. As we can see, our 
class-dependent weights $\beta$ are employed to extract a small\footnote{Our hypothesis is that the distribution of weights for a given document should be closer to the weights of its corresponding language variety, therefore, we use the most common descriptive statistics to measure this variability among language varieties.} --- but very discriminant --- number of features for each language variety.\footnote{Using the LDR a document is 
represented by a total set of features equal to 6 multiplied by the number of categories (the 5 language varieties), in our case 30 features. This is a considerable dimensionality 
reduction that may be helpful to deal with big data environments.} We note that the same process can be followed in order to represent a test document $d'\in D'$. We just need to 
use the $\beta$ matrix obtained with $D$ to index the document $d'$ by means of Eq.~\ref{eq:4}.

\begin{table}[H]
\begin{center}
\begin{tabular}{|l| l l l | }

    \hline
   \textbf{avg}$~~~~~~~~~~~~~~$ &     \multicolumn{3}{p{9.8cm}|}{
   	The average weight of a document is calculated as the sum of weights \textit{W(t,c)} of its terms divided by the total number of vocabulary terms of the document.
  }\\
     \hline
\textbf{std}$~~~~~~~~~~~~~~$ &     \multicolumn{3}{p{9.8cm}|}{
   	The standard deviation of the weight of a document is calculated as the root square of the sum of all the weights \textit{W(t,c)} minus the average.
  }\\
       \hline
\textbf{min}$~~~~~~~~~~~~~~$ &     \multicolumn{3}{p{9.8cm}|}{
   	The minimum weight of a document is the lowest term weight \textit{W(t,c)} found in the document.
  }\\
       \hline
\textbf{max}$~~~~~~~~~~~~~~$ &     \multicolumn{3}{p{9.8cm}|}{
   	The maximum weight of a document is the highest term weight \textit{W(t,c)} found in the document.
  }\\
       \hline
\textbf{prob}$~~~~~~~~~~~~~~$ &     \multicolumn{3}{p{9.8cm}|}{
   	The overall weight of a document is the sum of weights \textit{W(t,c)} of the terms of the document divided by the total number of terms of the document.
  }\\
       \hline
\textbf{prop}$~~~~~~~~~~~~~~$ &     \multicolumn{3}{p{9.8cm}|}{
   	The proportion between the number of vocabulary terms of the document and the total number of terms of the document.
  }\\
     \hline
\end{tabular}
\end{center}
\caption{Set of features for each category (language variety) used in Equation~\ref{eq:5}.}
\label{tab:features}
\end{table}

\vspace{-0.5cm}
\vspace{-0.5cm}
\vspace{-0.5cm}
\section{Evaluation Framework}

In this section, we describe the corpus and the alternative representations that we employ in this work.

\subsection{HispaBlogs Corpus}

We have created the HispaBlogs dataset\footnote{The HispaBlogs dataset was collected by experts on social media from the Autoritas Consulting company (\url{http://www.autoritas.net}). Autoritas experts in the different countries selected popular bloggers related to politics, online marketing, technology or trends. The HispaBlogs dataset is publicly available at: \url{https://github.com/autoritas/RD-Lab/tree/master/data/HispaBlogs}} by collecting posts from Spanish 
blogs from five different countries: Argentina, Chile, Mexico, Peru and Spain. For each country, there are 450 and 200 blogs respectively for training and test, ensuring that each author appears only in one set. Each blog contains 
at least 10 posts. The total number of blogs is 2,250 and 1,000 respectively. Statistics of the number of words are shown in Table~\ref{tab:hispablogs}.


\subsection{Alternative representations}

We are interested in investigating the impact of the proposed representation and compare its performance with state-of-the-art representations based on $n$-grams and with two 
approaches based on the recent and popular distributed representations of words by means of the continuous Skip-gram model \cite{franco2015}.

\begin{table}[H]
\begin{center}
\begin{tabular}{lcccccc}
\toprule
\multirow{2}{*}{\bf Language Variety} & \multicolumn{2}{@{}c}{\bf{\# Blogs/authors}} & \multicolumn{2}{@{}c}{\bf{\# Words}}  & \multicolumn{2}{@{}c}{\bf{\# Words per post}} \\ 
& \bf Training & \bf Test & \bf Training & \bf Test & \bf Training & \bf Test  \\ 
\midrule
AR - Argentina 	& 450 & 200 & 1,408,103 & 590,583 & 371 448 & 385 849\\
CL - Chile  		& 450 & 200 & 1,081,478 & 298,386 & 313 465 & 225 597 \\
ES - Spain 		& 450 & 200 & 1,376,478 &  620,778 & 360 426 & 395 765 \\
MX - Mexico 		& 450 & 200 & 1,697,091  & 618,502 & 437 513 & 392 894 \\
PE - Peru 		& 450 & 200 & 1,602,195  & 373,262 & 410 466 & 257 627 \\
\midrule
TOTAL & 2,250 & 1,000 & 7,164,935 & 2,501,511 & 380 466 & 334 764 \\
\toprule
\end{tabular}
\end{center}
\caption{\label{tab:hispablogs} Number of posts, words and words per post (average and standard deviation)\hspace{\textwidth} per language variety. }
\end{table}

\vspace{-0.5cm}
\vspace{-0.5cm}
\vspace{-0.5cm}
\subsubsection{State-of-the art representations}

State-of-the-art representations are mainly based on $n$-grams models, hence we tested character and word based ones, besides word with \textit{tf-idf} weights. For each of them, 
we iterated $n$ from 1 to 10 and selected 1,000, 5,000 and 10,000 most frequent grams. The best results were obtained with the 10,000 most frequent BOW, character $4$-grams and 
\textit{tf-idf} $2$-grams. Therefore, we will use them in the evaluation.

\vspace{-0.5cm}
\subsubsection{Distributed representations}


Due to the increasing popularity of the distributed representations \cite{hinton1986distributed}, we used the continuous Skip-gram model to generate distributed representations of 
words (e.g. $n$-dimensional vectors), with further refinements in order to use them with documents. The continuous Skip-gram 
model~\cite{mikolov2013efficient,mikolov2013distributed} is an iterative algorithm which attempts to maximize the classification of the context surrounding a word. Formally, given 
a word $w(t)$, and its surrounding words $w(t-c),~w(t-c+1),...,~w(t+c)$ inside a window of size $2c+1$, the training objective is to maximize the average of the log probability 
shown in Equation~\ref{eq:1}:
 
\begin{equation}
\frac{1}{T} \displaystyle\sum_{t=1}^T \displaystyle\sum_{-c \leq j \leq c,j \neq 0} \log p(w_{t+j}|w_t)
 \label{eq:1}
\end{equation}

To estimate $p(w_{t+j}|w_t)$ we used negative sampling~\cite{mikolov2013distributed} that is a simplified version of the Noise Contrastive Estimation 
(NCE)~\cite{gutmann2012noise,mnih2012fast} which is only concerned with preserving vector quality in the context of Skip-gram learning. The basic idea is to use logistic 
regression 
to distinguish the target word $W_O$ from draws from a noise distribution $P_n(w)$, having $k$ negative samples for each word.  Formally, the negative sampling estimates 
$p(w_O|w_I)$ following Equation~\ref{eq:2b}:
 
 \begin{equation}
\log\sigma(v'_{w_O}{}^T v_{w_I}) + \displaystyle\sum_{i=1}^k \mathbb{E}_{w_i}\sim P_n(w) \bigg[\log\sigma(-v'_{w_i}{}^T v_{w_I}) \bigg]
\label{eq:2b}
\end{equation}
 
 \noindent where $\sigma(x)=1/(1+\exp(-x))$. The experimental results in~\cite{mikolov2013distributed} show that this function obtains better results at the semantic level than 
hierarchical softmax \cite{goodman:2001} and NCE.   

In order to combine the word vectors to represent a complete sentence we used two approaches. First, given a list of word vectors $(w_1,w_2,...,w_n)$ 
belonging to a document, we generated a vector representation $v$ of its content by estimating the average of their dimensions: $v=n^{-1}\sum_{i=1}^n w_i$. We call this 
representation Skip-gram in the evaluation. In addition, we used Sentence vectors (SenVec)~\cite{le2014distributed}, a variant that follows
 Skip-gram architecture to train a special vector $sv$ representing the sentence. Basically, before each context window movement, SenVec uses a special vector $sv$ in place 
of $w(t)$ with the objective of maximizing the classification of the surrounding words. In consequence, $sv$ will be a distributed vector of the complete sentence.

Following state-of-the-art approach~\cite{le2014distributed}, in the evaluation we used a logistic classifier for both SenVec and Skip-gram approaches.\footnote{We used 
300-dimensional vectors, context windows of size 10, and 20 negative words for each sample.  We preprocessed the text with word lowercase, tokenization,
 removing the words of length one, and with phrase detection using word2vec tools:\\\url{https://code.google.com/p/word2vec/} }

\section{Experimental Results}

In this section we show experimental results obtained with the machine learning algorithms that best solve the problem with the proposed representation, the impact of the 
preprocessing on the performance, the obtained results in comparison with the ones obtained with state-of-the-art and distributed representations, the error analysis that provides 
useful insights to better understand differences among languages, a depth analysis on the contribution of the different features and a cost analysis that highlights the 
suitability 
of LDR for a big data scenario.

\subsection{Machine learning algorithms comparison}

We tested several machine learning algorithms\footnote{\url{http://www.cs.waikato.ac.nz/ml/weka/}} with the aim at selecting the one that best solves the task. As can be seen in Table \ref{tab:algorithms}, Multiclass 
Classifier\footnote{We used SVM with default parameters and exhaustive correction code to transform the multiclass problem into a binary one.} obtains the best result (results in 
the rest of the paper refer to Multiclass Classifier). We carried out a statistical test of significance 
with respect to the next two systems with the highest performance: SVM (\begin{math}z_{0.05} 0, 880 < 1, 960\end{math}) and LogitBoost (\begin{math}z_{0.05} = 1, 983 > 1, 960\end{math}).

\begin{table}[H]
\footnotesize
\begin{center}
  \scalebox{1}{
      \begin{tabular}{lc||lc||lc}
	  \hline \textbf{Algorithm} & \textbf{Accuracy} & \textbf{Algorithm} & \textbf{Accuracy}  & \textbf{Algorithm} & \textbf{Accuracy} \\ \hline
	  
	  Multiclass Classifier & \textbf{71.1} & Rotation Forest &  66.6 &  Multilayer Perceptron & 62.5 \\
	  SVM &  69.3 & Bagging & 66.5 &  Simple Cart &  61.9 \\
	  LogitBoost & 67.0 & Random Forest &  66.1 &  J48 &  59.3 \\
	  Simple Logistic &  66.8 &  Naive Bayes &  64.1 &  BayesNet & 52.2 \\	  		
	  \hline
      \end{tabular}
      }
\end{center}
  \caption{ \label{tab:algorithms} Accuracy results with different machine learning algorithms.} 
\end{table}
\vspace{-0.5cm}
\vspace{-0.5cm}
\subsection{Preprocessing impact}

\begin{figure}[H]
\centering
\includegraphics[width=1\linewidth]{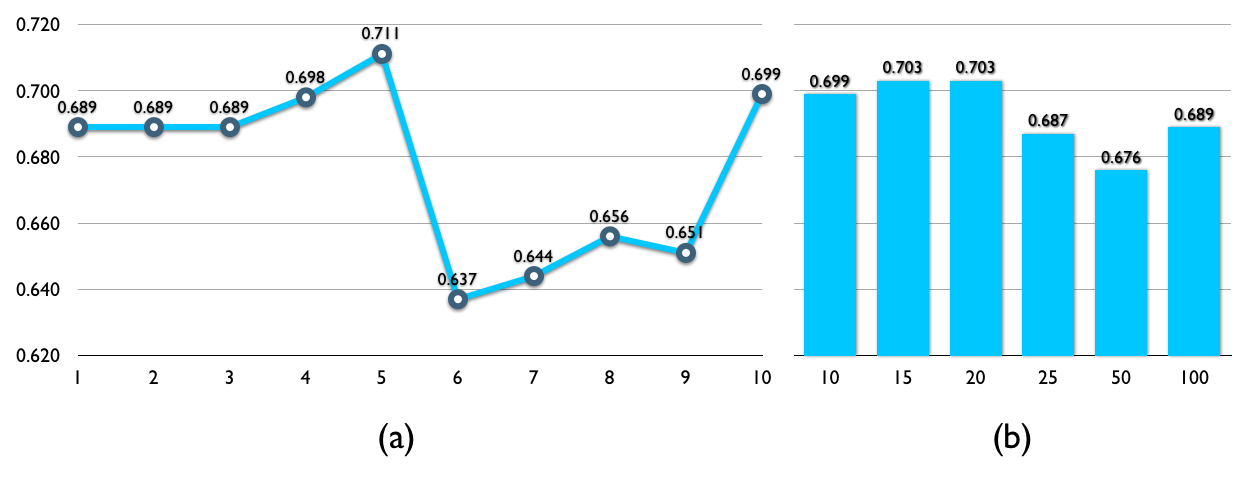}
\caption{Accuracy obtained after removing words with frequency equal or lower than $n$. \hspace{\textwidth}   (a) Continuous scale. (b) Non-continuous scale.}
  \label{fig:preprocessing}
\end{figure}

The proposed representation aims at using the whole vocabulary to obtain the weights of its terms. Social media texts may have noise and inadequately written words. Moreover, some 
of these words may be used only by few authors. With the aim at investigating their effect in the classification, we carried out a preprocessing step to remove words that appear 
less than $n$ times in the corpus, iterating $n$ between 1 and 100. In Figure \ref{fig:preprocessing} the corresponding accuracies are shown. In the left part of the figure 
(\textit{a}), results for $n$ between 1 and 10 are shown in a continuous scale. In the right part (\textit{b}), values from 10 to 100 are shown in a non-continuous scale. As can 
be 
seen, the best result was obtained with $n$ equal to 5, with an accuracy of 71.1\%. As it was expected, the proposed representation takes advantage from the whole vocabulary, 
although it is recommendable to remove words with very few occurrences that may alter the results. We show examples of those infrequent words in
Table~\ref{tab:resultsB}. 

\begin{table}[!ht]
\begin{center}
      \begin{tabular}{|l|l|l|}
	  \hline 
	  \textbf{\# occurrences = 1} & \textbf{\# occurrences =  2} & \textbf{\# occurrences =  3}    \\ 
	  \hline
	  	aaaaaaaah &  aaaaa & aaaa\\
	  	aaaaaaaarrrgh &  aaaayyy & aaaaaaaaae \\
	  	aaaaaaggghhhhh &  aaavyt & aaaaaaaacu \\
	  	aaaah & aach & aantofagastina \\
	  	aaaahhhh & aachen & \~nirripil \\	
	  	
	  \hline
      \end{tabular}
\end{center}
  \caption{ \label{tab:resultsB} Very infrequent words.} 
\end{table}

\begin{figure}[!ht]
\centering
\includegraphics[width=1\linewidth]{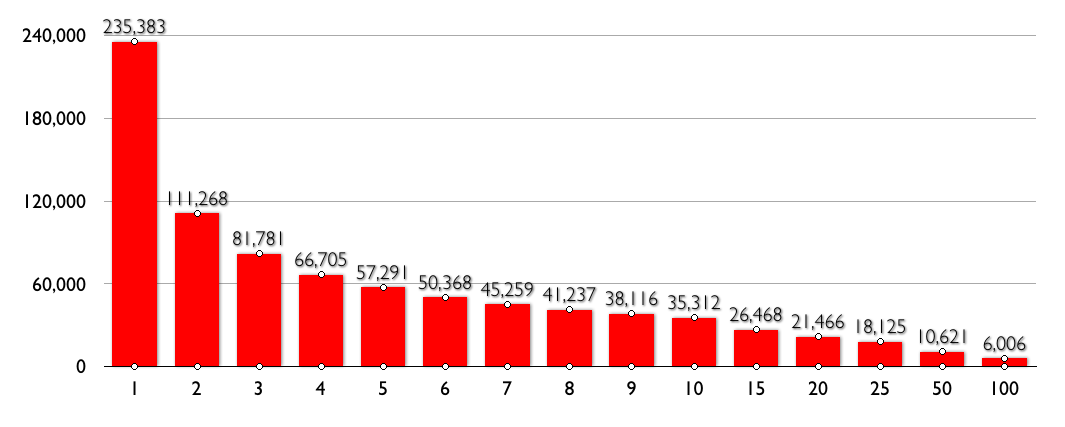}
\caption{Number of words after removing those with frequency equal or lower than $n$.}
  \label{fig:evol}
\end{figure}

In Figure \ref{fig:evol}, when analysing the evolution of the number of remaining words in function of the value of $n$, we can see a high number of words with very low frequency 
of occurrence. These words may introduce a high amount of noise in our LDR weight estimation. In addition, removing these words may be also beneficial in order to reduce the 
processing time needed to obtain the representation. This fact has special relevance for improving the performance in big data environments.


\subsection{Language variety identification results}

In Table \ref{tab:results}  we show the results obtained by the described representations employing the Multiclass Classifier. As can be appreciated, the proposed low 
dimensionality representation improves more than 35\% the results obtained with the state-of-the-art representations. BOW obtains slightly better results than character $4$-grams, 
and both of them improve significantly the ones obtained with \textit{tf-idf} $2$-grams. Instead of selecting the most frequent $n$-grams, our approach takes advantage from the 
whole vocabulary and assigns higher weights to the most discriminative words for the different language varieties as shown in Equation \ref{eq:3}.

\vspace{-0.5cm}
\begin{table}[H]
\begin{center}
  \scalebox{1}{
      \begin{tabular}{lcc}
	  \hline \textbf{Representation} & \textbf{Accuracy} & \textbf{\# Features} \\ \hline
	  Skip-gram & 0.722 & 300 \\
	   LDR & \textbf{0.711} & \textbf{30} \\
	   SenVec & 0.708 & 300 \\
	   BOW & 0.527 & 10,000 \\
	   Char. $4$-grams & 0.515 & 10,000 \\
	   \textit{tf-idf} $2$-grams & 0.393 & 10,000 \\
	   Random baseline & 0.200 & - \\
	  \hline
      \end{tabular}
      }
\end{center}
  \caption{ \label{tab:results}  Accuracy results in language variety identification and number of features for each representation.} 
\end{table}

\vspace{-0.5cm}
\vspace{-0.5cm}
We highlight that our LDR obtains competitive results compared with the use of distributed representations. Concretely, 
there is no 
significant difference among them (Skip-gram \begin{math}z_{0.05} = 0,5457 < 1,960\end{math} and SenVec\begin{math}z_{0.05} = 0,7095 < 1,960\end{math}). In addition, our proposal 
reduces considerably the dimensionality of one order of magnitude as shown in Table \ref{tab:results}.


\subsection{Error analysis}

We aim at analysing the error of LDR to better understand which varieties are the most difficult to discriminate. As can be seen in Table \ref{tab:confusion}, the 
Spanish variety is the easiest to discriminate. However, one of the highest confusions occurs from Argentinian to Spanish. Mexican and Spanish were considerably confused with 
Argentinian too. Finally, the highest confusion occurs from Peruvian to Chilean, although the lowest average confusion occurs with Peruvian. In general, Latin American varieties 
are closer to each other and it is more difficult to differentiate among them. Language evolves over time. It is logical that language varieties of nearby countries --- as the Latin American ones --- evolved in a more similar manner that the Spanish variety. It is also logical that even more language variety similarities are shared across neighbour countries, e.g. Chilean compared with Peruvian and Argentinian.

\vspace{-0.5cm}
\begin{minipage}[c]{.30\textwidth}
\begin{table}[H]
\begin{center}

\begin{tabular}{|c|rrrrr|}
\hline \multicolumn{1}{|c|}{} &\multicolumn{5}{|c|}{Clasified as} \\ \cline{2-6}
\multicolumn{1}{|c|}{Variety} & AR & CL & ES & MX & PE\\ \hline
AR & 143 & 16 & 22 & 8 & 11 \\
CL & 17 & 151 & 11 & 11 & 10 \\
ES & 20 & 13 & 154 & 7 & 6 \\
MX & 20 & 18 & 18 & 131 & 13 \\
PE & 16 & 28 & 12 & 12 & 132 \\
\hline
\end{tabular}

\end{center}
\caption{ \label{tab:confusion} Confusion matrix of the 5-class classification.}
\end{table} 
\end{minipage}
\hfill
\begin{minipage}[c]{.60\textwidth}
\begin{figure}[H]
\centering
\includegraphics[width=.7\linewidth]{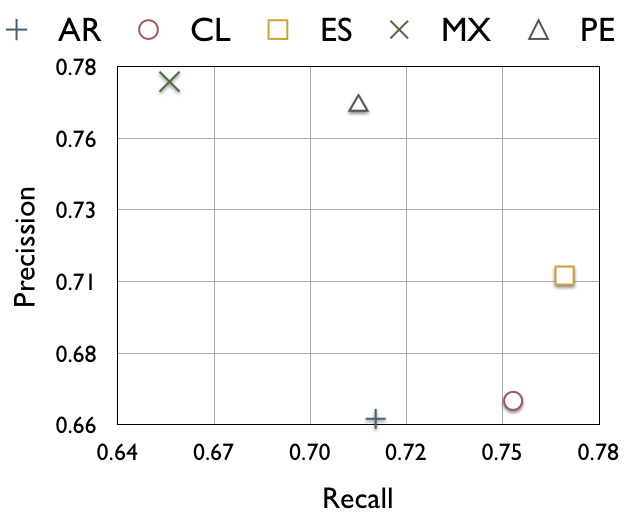}
\caption{F1 values for identification as the corresponding language variety vs. others.}
  \label{fig:quadrant}
\end{figure}
\end{minipage}\\

In Figure \ref{fig:quadrant} we show the precision and recall values for the identification of each variety. As can be seen, Spain and Chile have the highest recall so that texts 
written in these varieties may have less probability to be misclassified as other varieties. Nevertheless, the highest precisions are obtained for Mexico and Peru, implying that texts 
written in such varieties may be easier to discriminate.





\subsection{Most discriminating features}

In Table \ref{tab:mostdiscriminantfeatures} we show the most discriminant features. The features are sorted by their information gain (IG).  As can be seen, the highest gain is 
obtained by average, maximum and minimum, and standard deviation. On the other hand, probability and proportionality features has low information gain. 

\begin{table}[!ht]
\begin{center}
      \begin{tabular}{|lc| |lc| |lc|}
	  \hline \textbf{Attribute} & \textbf{IG} & \textbf{Attribute} & \textbf{IG} & \textbf{Attribute} & \textbf{IG} \\ \hline
	  	PE-avg & 0.680 $\pm$ 0.006 & ES-std & 0.497 $\pm$ 0.008 & PE-prob & 0.152 $\pm$ 0.005 \\
		AR-avg & 0.675 $\pm$ 0.005 & CL-max & 0.496 $\pm$ 0.005 & MX-prob & 0.151 $\pm$ 0.005 \\
		MX-max & 0.601 $\pm$ 0.005 & CL-std & 0.495 $\pm$ 0.007 & ES-prob & 0.130 $\pm$ 0.011 \\
		PE-max & 0.600 $\pm$ 0.009 & MX-std & 0.493 $\pm$ 0.007 & AR-prob & 0.127 $\pm$ 0.006 \\
		ES-min & 0.595 $\pm$ 0.033 & CL-min & 0.486 $\pm$ 0.013 & AR-prop & 0.116 $\pm$ 0.005 \\
		ES-avg & 0.584 $\pm$ 0.004 & AR-std & 0.485 $\pm$ 0.005 & MX-prop & 0.113 $\pm$ 0.006 \\
		MX-avg & 0.577 $\pm$ 0.008 & PE-std & 0.483 $\pm$ 0.012 & PE-prop & 0.112 $\pm$ 0.005 \\
		ES-max & 0.564 $\pm$ 0.007 & AR-min & 0.463 $\pm$ 0.012 & ES-prop & 0.110 $\pm$ 0.007 \\
		AR-max & 0.550 $\pm$ 0.007 & CL-avg & 0.455 $\pm$ 0.008 & CL-prop & 0.101 $\pm$ 0.005 \\
		MX-min & 0.513 $\pm$ 0.027 & PE-min & 0.369 $\pm$ 0.019 & CL-prob & 0.087 $\pm$ 0.010 \\
	  \hline
      \end{tabular}
\end{center}
  \caption{ \label{tab:mostdiscriminantfeatures} Features sorted by information gain.} 
\end{table}

\begin{figure}[H]
\centering
\includegraphics[width=1\linewidth]{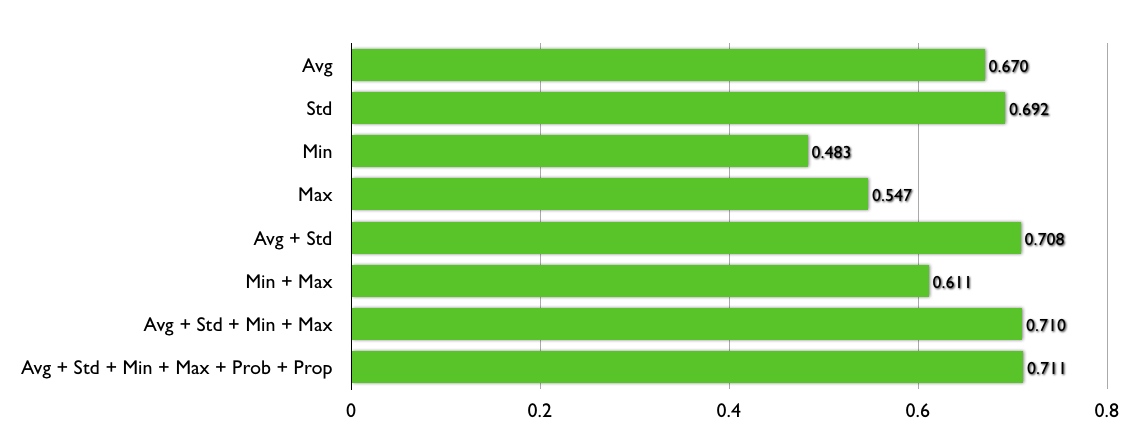}
\caption{Accuracy with different combinations of features.}
  \label{fig:mostdiscriminatingfeatures}
\end{figure}

We experimented with different sets of features and show the results in Figure \ref{fig:mostdiscriminatingfeatures}. As may be expected, average-based features obtain high 
accuracies (67.0\%). However, although features based on standard deviation have not the highest information gain, they  obtained the highest results individually (69.2\%), as 
well 
as their combination with average ones (70,8\%). Features based on minimum and maximum obtain low results (48.3\% and 54.7\% respectively), but in combination they obtain a 
significant increase (61.1\%). The combination of the previous features obtains almost the highest accuracy (71.0\%), equivalent to the accuracy obtained with probability and proportionality features (71.1\%).

\vspace{-0.5cm}
\vspace{-0.4cm}


\vspace{-0.9cm}
\subsection{Cost analysis}

We analyse the cost from two perspectives: \textit{i)} the complexity to the features; and \textit{ii)} the number of features needed to represent a document. Defining 
$l$ as the number of different language varieties, and $n$ the number of terms of the document to be classified, the cost of obtaining the features of Table \ref{tab:features} 
(average, minimum, maximum, probability and proportionality) is 
$O(l\cdot{n})$. Defining $m$ as the number of terms in the document that coincides with some term in the vocabulary, the cost of obtaining the standard deviation is 
$O(l\cdot{m})$. 
As the average is needed previously to the standard deviation calculation, the total cost is $O(l\cdot{n}) + O(l\cdot{m})$ that is equal to $O(max(l\cdot{n}, l\cdot{m})) = 
O(l\cdot{n})$. Since the number of terms in the vocabulary will always be equal or greater than the number of coincident terms (\begin{math}n \geq m\end{math}), and as the number 
of terms in the document will always be much higher than the number of language varieties (\begin{math}l<<n\end{math}), we can determine the cost as lineal with respect to the 
number of terms in the document $O(n)$. With respect to the number of features needed to represent a document, we showed in Table \ref{tab:results} the considerable reduction of the proposed low dimensionality 
representation.

\vspace{-0.2cm}
\subsection{Robustness}

In order to analyse the robustness of the low dimensionality representation to different languages, we experimented with the development set of the DSLCC 
corpus\footnote{\url{http://ttg.uni-saarland.de/lt4vardial2015/dsl.html}} from the Discriminating between Similar Languages task \cite{tan2014}. The corpus consists of 2,000 
sentences per language or variety, with between 20 and 100 tokens per sentence, obtained from news headers.  In Table~\ref{tab:robustness} we show the results obtained with the 
proposed representation and the two distributed representations, Skip-gram and SenVec. 
It is important to notice that, in general, when a particular representation improves for one language is at cost of the 
other one. We can conclude that the three representations obtained comparative results and support the robustness of the low dimensionality representation.

\vspace{-0.5cm}
\begin{table}[H]
  \begin{center}
  \scalebox{1}{
      \begin{tabular}{lccc}
	  \hline \textbf{Language} & \textbf{LDR} & \textbf{Skip-gram} & \textbf{SenVec} \\ \hline
		Bulgarian & 99.9 & 100 & 100  \\
		Macedonian & 99.9 & 100 & 100  \\
		\hline
		Spain Spanish & \textbf{84.7}  & 82.1 & \textbf{86.3}\\
		Argentina Spanish & 88.0  & \textbf{90.3} & 87.6 \\
		\hline
		Portugal Portuguese & \textbf{87.4}  & 83.2 & \textbf{90.0}\\
		Brazilian Portuguese & \textbf{90.0}  & \textbf{94.5} & 87.6\\
		\hline
		Bosnian &  \textbf{78.0}  & 80.3 & 74.4\\
		Croatian &  85.8  & 85.9 & 84.7\\
		Serbian & \textbf{86.4}  & 75.1 & 91.2 \\
		\hline
		Indonesian &  99.4  & 99.3 & 99.4\\
		Malay & 99.2  & 99.2 & \textbf{99.8}\\
		\hline
		Czech & 99.8  & 99.9 & 99.8\\
		Slovak & 99.3  & \textbf{100} & 99.3\\
		\hline
		Other languages &  99.9  & 99.8 & 99.8 \\

	  \hline
      \end{tabular}
      }
      \end{center}
  \caption{ \label{tab:robustness} Accuracy results in the development set of the DSLCC. The significance is marked in bold when some representation obtains significantly better results than the next best performing representation (e.g. results for SenVec in Portugal Portuguese are significantly higher than LDR, which at the same time are significantly higher than Skip-gram).} 
\end{table}

\vspace{-0.5cm}
\vspace{-0.5cm}
\vspace{-0.5cm}
\section{Conclusions}

In this work, we proposed the LDR low dimensionality representation for language variety identification. Experimental results outperformed traditional state-of-the-art 
representations and obtained competitive results compared with two distributed representation-based approaches that employed the popular continuous Skip-gram model.
The dimensionality reduction obtained by means of LDR is from thousands to only 6 features per language variety. This allows to deal with large collections in big data 
environments such as social media. Recently, we have applied LDR to the age and gender identification task obtaining competitive results with the best performing 
teams in the author profiling task at the PAN\footnote{\url{http://pan.webis.de}} Lab at CLEF.\footnote{\url{http://www.clef-innitiative.org}} As a future work, we plan to apply 
LDR to other author profiling tasks such as personality recognition.

\end{document}